\documentclass[conference]{IEEEtran}

\usepackage{amsmath,amsfonts,amssymb}
\usepackage{algorithmic}
\usepackage{array}
\usepackage[caption=false,font=normalsize,labelfont=sf,textfont=sf]{subfig}
\usepackage{textcomp}
\usepackage{stfloats}
\usepackage{url}
\usepackage{verbatim}
\usepackage{graphicx}
\usepackage{cite}

\usepackage{enumitem}
\usepackage{booktabs}
\usepackage{xcolor}
\usepackage[left=1.91cm, right = 1.91cm, top=2.54cm, bottom=1.91cm]{geometry}
\usepackage{multirow}
\usepackage{float}

\IEEEoverridecommandlockouts
\def\BibTeX{{\rm B\kern-.05em{\sc i\kern-.025em b}\kern-.08em
    T\kern-.1667em\lower.7ex\hbox{E}\kern-.125emX}}
\usepackage{balance}

\title{\LARGE 
DTEA: A Dual-Topology Elastic Actuator Enabling Real-Time Switching Between Series and Parallel Compliance
}
\author{
Vishal Ramesh$^{1}$, Aman Singh$^{2}$, and Shishir Kolathaya$^{2}$%
\thanks{This research is supported by ARTPARK at IISc.}%
\thanks{$^{1}$Vishal Ramesh is with the Department of Cyber-Physical Systems (CPS), Indian Institute of Science (IISc), Bengaluru, India, and the Indian Institute of Information Technology, Hyderabad, India.
{\tt\scriptsize vishal0905@berkeley.edu}}%
\thanks{$^{2}$Aman Singh and Shishir Kolathaya are with the Department of Cyber-Physical Systems (CPS), Indian Institute of Science (IISc), Bengaluru, India.
{\tt\scriptsize \{saman, shishirk\}@iisc.ac.in}}%
}

\begin{document}

\maketitle
\thispagestyle{empty}
\pagestyle{empty}


\begin{abstract}
Series and parallel elastic actuators offer complementary but mutually exclusive advantages, yet no existing actuator enables real-time transition between these topologies during operation. This paper presents a novel actuator design called the \textit{Dual-Topology Elastic Actuator (DTEA)}, which enables dynamic switching between SEA and PEA topologies during operation. 
A proof-of-concept prototype of the DTEA is developed to demonstrate the feasibility of the topology-switching mechanism. Experiments are conducted to evaluate the robustness and timing of the switching mechanism under operational conditions. The actuator successfully performed 324 topology-switching cycles under load without damage, demonstrating the robustness of the mechanism. The measured switching time between SEA and PEA modes is under $33.33$\,ms.
Additional experiments are conducted to characterize the static stiffness and disturbance rejection performance in both SEA and PEA modes. Static stiffness tests show that the PEA mode is $1.53\times$ stiffer than the SEA mode, with $K_{\text{SEA}} = 5.57 \pm 0.02$\,Nm/rad and $K_{\text{PEA}} = 8.54 \pm 0.02$\,Nm/rad. Disturbance rejection experiments show that the mean peak deflection in SEA mode is $2.26\times$ larger than in PEA mode ($5.2^\circ$ vs.\ $2.3^\circ$), while the mean settling time is $3.45\times$ longer ($1380$\,ms vs.\ $400$\,ms). The observed behaviors are consistent with the known characteristics of conventional SEA and PEA actuators, validating the functionality of both modes in the DTEA actuator.
\end{abstract}

\begin{IEEEkeywords}
Actuation and Joint Mechanisms; Mechanism Design; Engineering for Robotic Systems; Series Elastic Actuators; Parallel Elastic Actuators; Topology switching;
\end{IEEEkeywords}

\section{Introduction}
\label{sec:introduction}

Compliant actuation is central to robots that interact with unstructured environments and humans. The Series Elastic Actuator (SEA)~\cite{Pratt1995} places a spring between motor and load, reducing interface stiffness while enabling shock tolerance, low reflected inertia, force estimation from deflection, and energy storage. SEAs are widely used in exoskeletons~\cite{Yu2015} and legged robots~\cite{Hurst2010}. However, all load torque passes through the spring, causing continuous I$^2$R losses even in quasi-static postures.

The Parallel Elastic Actuator (PEA) addresses this limitation by placing the spring between the actuator housing and the output, directly coupling the elastic element in parallel with the motor. When tuned to the dominant static load, the spring passively compensates gravity, reducing energy consumption by up to 78--80\%~\cite{Verstraten2016, Haeufle2012}. This energy advantage has motivated PEA adoption in prosthetic ankles~\cite{Au2009}, bipedal robots~\cite{Mazumdar2017}, and exoskeletons~\cite{Wang2022}. However, the spring torque is rigidly coupled to the joint angle, meaning the motor must work against the spring for any motion away from equilibrium, making PEAs ill-suited for large range-of-motion tasks.

\begin{figure}[t]
    \centering
    \includegraphics[width=0.85\columnwidth]{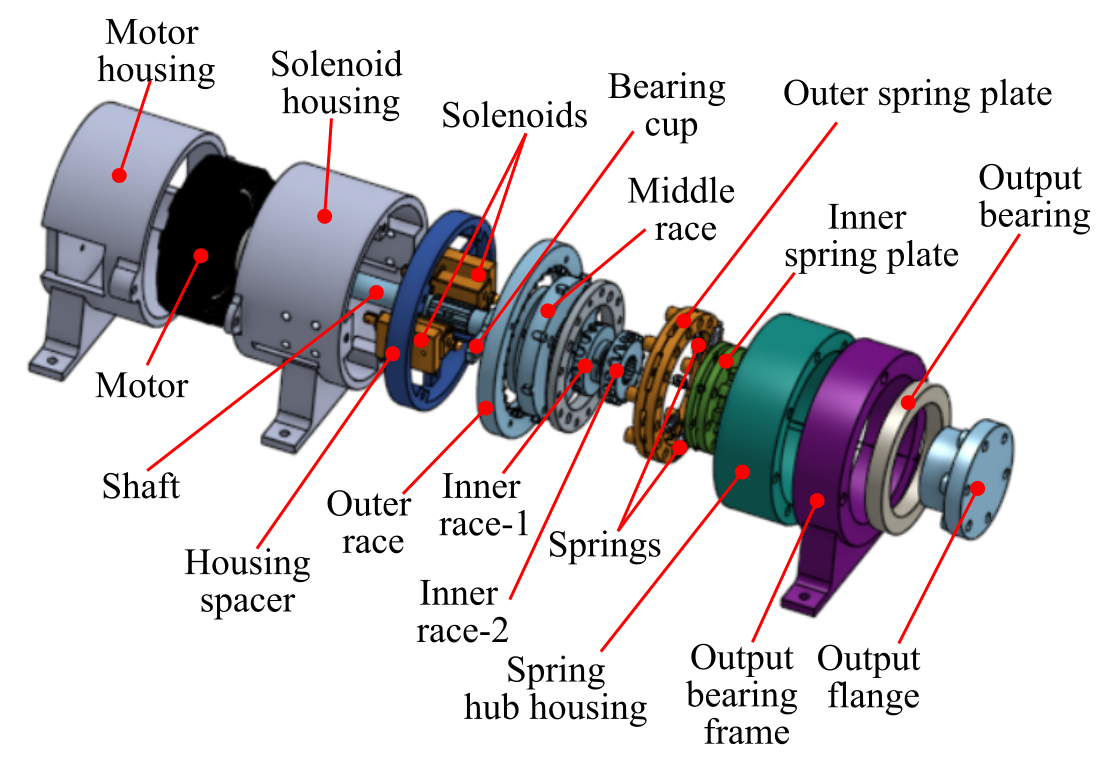}
    \caption{\textbf{DTEA:} Exploded view of Dual-Topology Elastic Actuator prototype, an actuator that switches between SEA and PEA modes during operation.}
    \label{fig:exploded}
\end{figure}

The question of which topology is energetically superior has no universal answer. Verstraten et al.~\cite{Verstraten2016} demonstrated that the crossover between SEA and PEA efficiency depends on gravitational offset and spring tuning, Beckerle et al.~\cite{Beckerle2017} showed the energy-optimal topology varies with operating position, and Yesilevskiy et al.~\cite{Yesilevskiy2018} found task-dependent advantages for both topologies in hopping monopeds. The consistent conclusion is that the optimal choice depends on the operating regime, which in legged locomotion changes within a single stride~\cite{Geyer2006}. Prior work has sought to combine these benefits through a variety of variable impedance architectures~\cite{Vanderborght2013}, including clutched parallel springs~\cite{Haeufle2012, Plooij2016, Geeroms2017}, clutchable series elastic actuators that lock the motor during elastic phases~\cite{Rouse2014}, fixed series-parallel combinations~\cite{Mathijssen2015, Roozing2021}, variable stiffness mechanisms~\cite{Wolf2016, Toubar2022}, and mode-switchable gear trains~\cite{Zhu2025}. However, these approaches modify stiffness magnitude, spring engagement timing, or gear configuration within a fixed topology. None switches the elastic element grounding path itself between the series and parallel topologies. This distinction is particularly relevant for legged robots and lower-limb exoskeletons, where heel strike favors SEA compliance and push-off favors PEA efficiency within a single stride. Table~\ref{tab:comparison} summarizes this landscape.

The key observation is that the elastic element in both SEA and PEA topologies is mechanically identical. What differs is where the spring connects. In an SEA, the spring connects the motor to the load. In a PEA, the same spring connects the actuator housing to the load, while the motor is directly coupled to the load. If one end of a single elastic element can be redirected between the motor shaft and actuator housing in real time, the same physical spring can function as a series element in one configuration and a parallel element in the other. In this paper, we present the Dual-Topology Elastic Actuator (DTEA), which realizes this concept.
The contributions of this work are:
\begin{itemize}
\item We propose a \textbf{Dual Topology Elastic Actuator (DTEA)} that switches between \textit{Series Elastic Actuation (SEA)} and \textit{Parallel Elastic Actuation (PEA)} during operation. The design allows a single elastic element to be reconfigured between both architectures without interrupting torque transmission.

\item We build a proof-of-concept prototype and evaluate topology switching under cyclic loading to test the robustness of the mechanism. The actuator transitions between the two configurations within ${<}\,33.33$\,ms.

\item We experimentally evaluate the actuator using \textit{static stiffness characterization} and \textit{disturbance rejection under position control}. The results are compared with known characteristics of conventional SEA and PEA systems, showing that the DTEA reproduces their expected behavior.

\end{itemize}

\begin{table*}[t]
\centering
\caption{Comparison of Compliant Actuator Designs}
\label{tab:comparison}
\renewcommand{\arraystretch}{1.2}
\begin{tabular}{l c c c c}
\hline
\textbf{Approach} & \textbf{Topology} & \textbf{Topology Switching} & \textbf{Stiffness Modulation} & \textbf{Energy Reduction} \\
\hline
SEA~\cite{Pratt1995} & Series & No & No & --- \\
PEA~\cite{Verstraten2016} & Parallel & No & No & 78--80\% \\
CPEA~\cite{Haeufle2012} & Parallel (clutched) & No & No (engage/disengage) & 80\% \\
BIC-PEA~\cite{Plooij2016} & Parallel (clutched) & No & No (bidirectional) & 65\% \\
CSEA~\cite{Rouse2014} & Series (clutched) & No & No (lock/unlock motor) & 70\%$^\dagger$ \\
SPEA~\cite{Roozing2021} & Series + Parallel (fixed) & No & Discrete (gear-based) & 65--75\% \\
VSA~\cite{Wolf2016} & Single (variable $K$) & No & Continuous/Discrete & --- \\
SEADAS~\cite{Toubar2022} & Series (variable $K$) & No & Discrete & --- \\
Mode-switchable~\cite{Zhu2025} & Gear reconfiguration & No$^*$ & Implicit & --- \\
\textbf{DTEA (this work)} & \textbf{Series $\boldsymbol{\leftrightarrow}$ Parallel} & \textbf{Yes ($\boldsymbol{<}$\,33.33\,ms)} & \textbf{Inherent ($\boldsymbol{2.08\times}$)} & \textbf{---}$^{\ddagger}$ \\
\hline
\multicolumn{5}{l}{\footnotesize $^\dagger$Simulation only. \quad $^*$Switches gear train, not elastic element grounding path. \quad $^\ddagger$Energy reduction validation planned as future work.}
\end{tabular}
\end{table*}

\section{Analytical Framework}
\label{sec:dynamics}

SEA and PEA offer complementary advantages, with the preferred topology depending on operating regime~\cite{Verstraten2016, Beckerle2017}. This section presents their standard dynamic models. SEA consists of two inertias coupled by a spring, while PEA forms a single rigid body with a grounded spring. This structural difference leads to distinct stiffness and dynamic behavior under identical control, validated experimentally in Section~\ref{sec:experiments}.
Throughout, we consider a direct-drive rotary 
joint~\cite{Wensing2017} with motor inertia~$J_m$, output 
inertia~$J_o$, spring stiffness~$K_s$, motor 
torque~$\tau_m$, and external load torque~$\tau_\text{ext}$.

\subsection{Equations of Motion}
\label{sec:eom}

In SEA mode, the spring sits between the motor and the 
output. The motor pushes the spring, and the spring pushes 
the load. The two can move independently. The equations of 
motion are~\cite{Pratt1995}
\begin{equation}
  J_m \ddot{\theta}_m = \tau_m - K_s(\theta_m - \theta_o)
  \label{eq:sea_motor}
\end{equation}
\begin{equation}
  J_o \ddot{\theta}_o = K_s(\theta_m - \theta_o) 
  - \tau_\text{ext}
  \label{eq:sea_output}
\end{equation}

At steady state, summing~\eqref{eq:sea_motor} 
and~\eqref{eq:sea_output} yields 
$\tau_m = \tau_\text{ext}$: the motor must supply the full 
external load regardless of the 
spring~\cite{Pratt1995, Verstraten2016}. This is the 
fundamental energy penalty of series elasticity.

In PEA mode, the motor and load are rigidly coupled, giving 
$\theta_m = \theta_o \triangleq \theta$. The spring 
connects the load to the stationary housing rather than to 
the motor~\cite{Grimmer2012}:
\begin{equation}
  (J_m + J_o)\,\ddot{\theta} = \tau_m - K_s\,\theta 
  - \tau_\text{ext}
  \label{eq:pea_combined}
\end{equation}

\noindent where $\theta$ is measured relative to the 
spring's zero-torque equilibrium. At steady state, $\tau_m = \tau_\text{ext} + K_s\,\theta$. If the spring is tuned such that $K_s\,\theta_\text{eq} \approx -\tau_\text{ext}(\theta_\text{eq})$, the required motor torque approaches zero, enabling gravity compensation and up to 78--80\% energy reduction~\cite{Verstraten2016, Grimmer2012}.

\subsection{Consequences of the Structural Difference}
\label{sec:consequences}

The difference between two angle variables 
in~\eqref{eq:sea_motor}--\eqref{eq:sea_output} and one 
in~\eqref{eq:pea_combined} has two well-established 
consequences.

First, the two-mass SEA model can oscillate between motor and load through the spring, and a motor-side position controller cannot suppress this because it does not observe the load~\cite{Calanca2016}. The single-mass PEA model cannot 
oscillate in this way because there is no spring between 
motor and load~\cite{Grimmer2012}. This is confirmed by 
the disturbance rejection experiment in 
Section~\ref{sec:exp_disturbance}.

Second, the SEA energy penalty ($\tau_m = \tau_\text{ext}$ 
at steady state) and the PEA energy advantage 
($\tau_m \approx 0$ at the tuned equilibrium) mean that 
the preferred topology depends on the operating 
condition~\cite{Verstraten2016, Beckerle2017, 
Yesilevskiy2018}. The dynamic switching experiment in 
Section~\ref{sec:exp_dynamic_switching} confirms this 
with a $4.93\times$ current reduction.

These properties have been studied extensively in 
isolation. To our knowledge, no prior work has compared them on the same 
actuator because no actuator has been capable of switching 
between both topologies. The DTEA mechanism described in 
Section~\ref{sec:mech_design} enables this comparison, 
and the experiments in Section~\ref{sec:experiments} 
provide the first side-by-side validation.

\subsection{Open-Loop Stiffness}
\label{sec:ol_stiffness}
Open-loop stiffness in each SEA and PEA mode is defined as the motor torque required to produce a unit angular displacement when the actuator output is mechanically fixed to the ground. This measures the intrinsic mechanical stiffness of the actuator in each topology. To compare the two modes, we consider torque control with the actuator output locked ($\theta_o = 0$). This isolates mechanical stiffness from controller effects and matches the protocol in Section~\ref{sec:exp_stiffness}.

In SEA, with the output locked, any motor rotation directly 
stretches the spring. From~\eqref{eq:sea_motor} at steady 
state~\cite{Pratt1995},
\begin{equation}
  \tau_m = K_s\,\theta_m
  \label{eq:sea_ol}
\end{equation}

The apparent stiffness is 
$K_\text{SEA}^\text{OL} = K_s$.

In PEA, the rigid coupling constrains 
$\theta_m = \theta_o$. With the output locked, the motor 
cannot rotate in an ideal rigid prototype, yielding infinite 
apparent stiffness. In the physical prototype, the 
3D-printed components flex slightly under load, producing a 
finite structural compliance $K_\text{struct}$:
\begin{equation}
  \tau_m = (K_s + K_\text{struct})\,\theta_m
  \label{eq:pea_ol}
\end{equation}

The apparent stiffness is 
$K_\text{PEA}^\text{OL} = K_s + K_\text{struct}$, where 
$K_\text{struct}$ is experimentally inferred as 
$K_\text{struct} = K_\text{PEA}^\text{OL} - K_s$. 

This value represents the combined effect of structural 
flexibility in the 3D-printed components, differences in 
spring engagement geometry between modes, and any friction 
asymmetry. Isolating individual contributions would require 
separate characterization of each source.

The ratio between topologies is
\begin{equation}
  \frac{K_\text{PEA}^\text{OL}}{K_\text{SEA}^\text{OL}} 
  = 1 + \frac{K_\text{struct}}{K_s}
  \label{eq:ol_ratio}
\end{equation}

From measured values in the deflection range where both 
modes remain linear (Section~\ref{sec:exp_stiffness}), 
$K_\text{SEA}^\text{OL} = 4.09$~Nm/rad and 
$K_\text{PEA}^\text{OL} = 8.49$~Nm/rad, giving a ratio 
of~2.08 and $K_\text{struct} \approx 1.08\,K_s$.


\section{Mechanical Design}
\label{sec:mech_design}

\begin{figure}[t]
    \centering
    \includegraphics[width=0.8\linewidth]{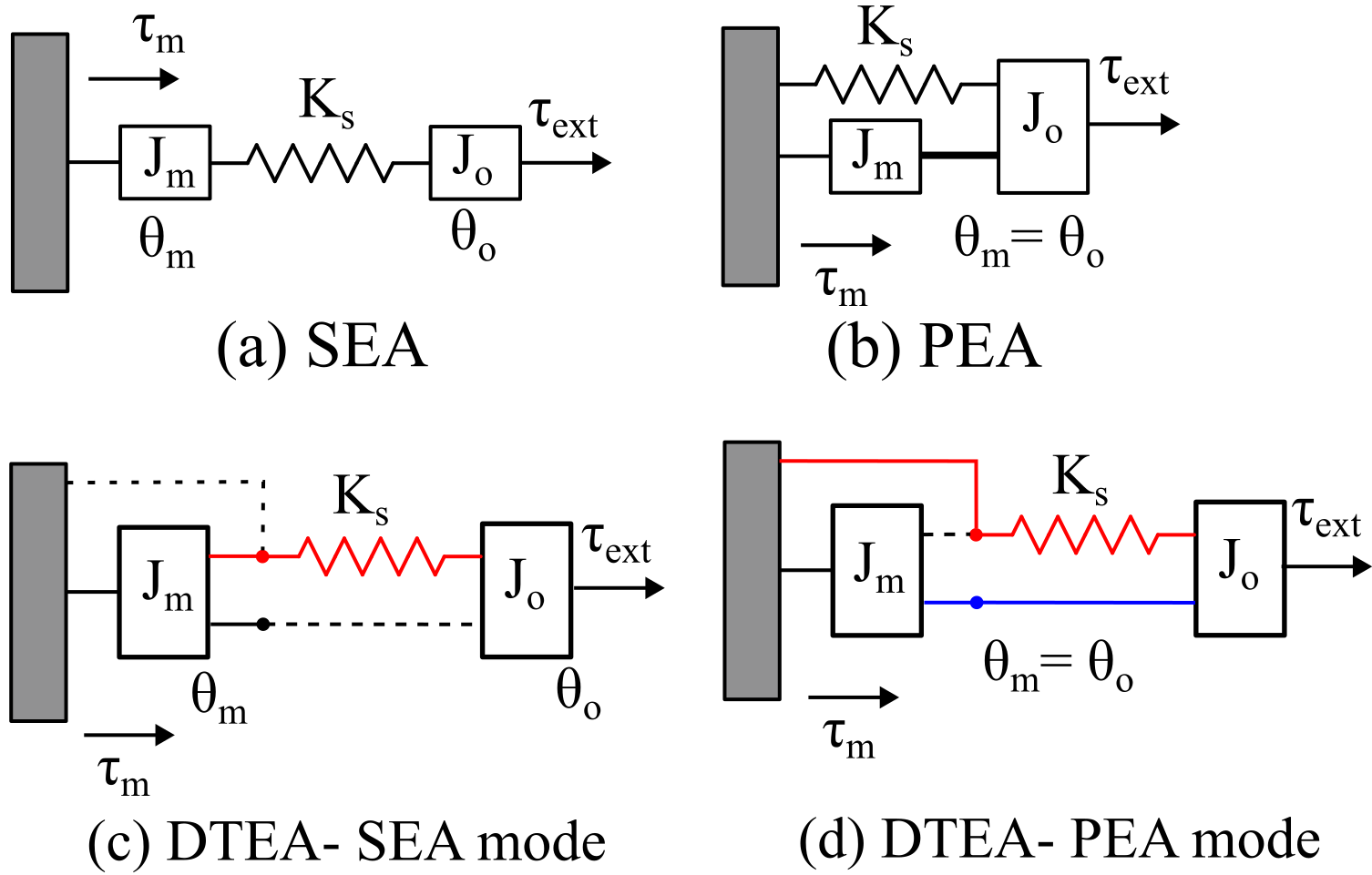}
    \caption{Spring-line schematics of standard (SEA \& PEA) and DTEA topologies. 
(a)~SEA: spring $K_s$ in series between motor inertia $J_m$ and load inertia $J_o$. 
(b)~PEA: rigid motor--load coupling with $K_s$ grounded to the housing. 
(c)~DTEA in SEA mode: spring engaged between motor shaft (in red) and load. 
(d)~DTEA in PEA mode: spring grounded to housing (in red) with rigid motor--load path (in blue). 
\textit{Note:} DTEA performs this topology switch with a single actuation.}
    \label{fig:spring_line_diagrams}
\end{figure}

The DTEA uses a single elastic element whose grounding point is switched between topologies. In SEA mode the spring connects the motor shaft to the load. In PEA mode the same spring is grounded to the housing while a rigid path couples the motor directly to the load. Fig.~\ref{fig:spring_line_diagrams} illustrates the topology change. Fig.~\ref{fig:exploded} shows the assembly and Fig.~\ref{fig:cross_section} shows the torque paths in both modes.

\subsection{System Architecture}
\label{sec:architecture}

The DTEA prototype is arranged coaxially along a central shaft and consists of five functional groups (Fig.~\ref{fig:exploded}): a direct-drive BLDC motor, a central shaft rigidly coupled to the motor, a three-race selector mechanism, a radial spring hub, and an output flange. Motor and transmission parameters are summarized in Table~\ref{tab:specs}. The direct-drive layout isolates topology-dependent effects from gear train dynamics~\cite{Wensing2017}.

The three-race selector switches the spring connection by translating a middle race axially between two engagement states (Section~\ref{sec:three_race}). The radial spring hub provides the elastic element shared by both modes (Section~\ref{sec:spring_hub}). The output flange connects to the inner spring plate through a spline interface that transmits torque while allowing the axial motion required for switching. Output angle $\theta_o$ is measured using an AS5047P absolute encoder mounted on the output side.

\subsection{Three-Race Selector Mechanism}
\label{sec:three_race}

\begin{figure*}[htbp]
    \centering
    \includegraphics[width=0.8\textwidth]{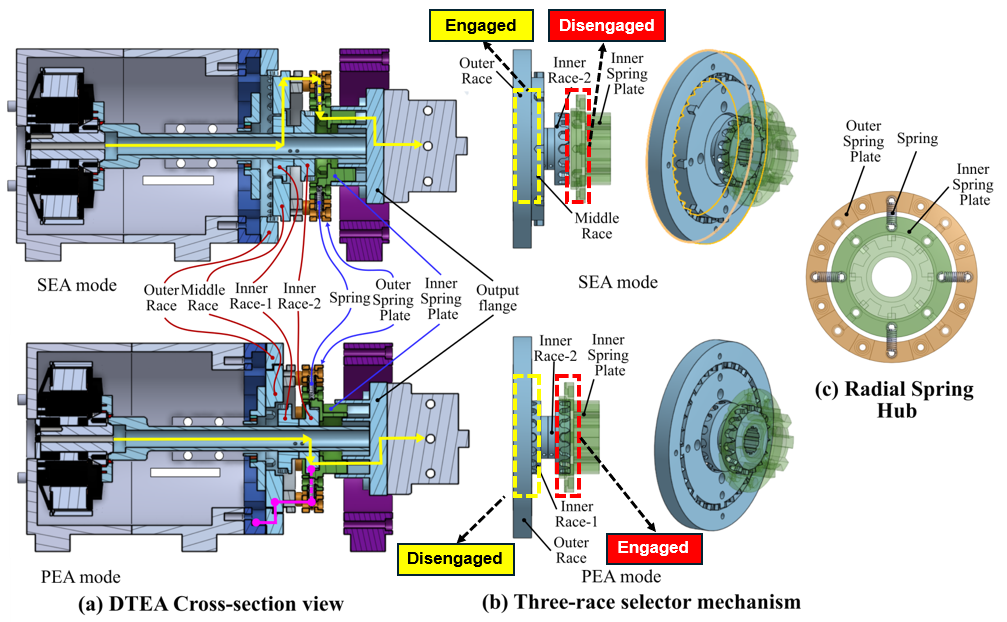}
    \caption{\textit{(a)}~Cross-sectional view of the DTEA in SEA mode (top) and PEA mode (bottom), where yellow arrows trace the torque transmission path. \textit{(b)}~Three-race selector mechanism in SEA mode (top) and PEA mode (bottom), showing engaged and disengaged interfaces between the outer race, middle race, Inner Race-1, Inner Race-2, and inner spring plate. \textit{(c)}~Radial spring hub showing the outer spring plate, inner spring plate, and four extension springs.}
    \label{fig:cross_section}
\end{figure*}

The topology-switching functionality of the DTEA is enabled by a three-race selector mechanism composed of three concentric elements: an outer race, a middle race, and a dual-part inner race (Fig.~\ref{fig:cross_section}(b)).

The \textit{outer race} is a hollow cylindrical structure fixed to the actuator housing and therefore grounded. Dog teeth on its inner circumference enable engagement with the middle race.

The \textit{middle race} is a cylindrical sleeve that slides axially on the actuator shaft to switch between SEA and PEA modes. It contains dog teeth on both its outer and inner circumferences, which engage the outer race and the inner race, respectively. A solenoid (25~N, $\approx 10$~mm stroke) translates it along the shaft to activate the desired interface. It is rigidly connected to the outer spring plate (Section~\ref{sec:spring_hub}).

The \textit{inner race} consists of two components: Inner Race--1 and Inner Race--2. Both are rigidly attached to the actuator shaft and rotate with the motor. Each component has dog teeth on its outer periphery. Inner Race--1 selectively engages with the middle race, while Inner Race--2 interfaces with the inner spring plate of the radial spring hub assembly (Section~\ref{sec:spring_hub}).

\subsection{Radial Spring Hub}
\label{sec:spring_hub}

The radial spring hub assembly provides the primary compliance of the actuator and consists of three components: an outer spring plate, an inner spring plate, and extension springs connecting the two plates.

The \textit{outer spring plate} is a circular plate with a central bore that rotates about the actuator shaft. It is rigidly connected to the middle race of the selector mechanism in all operating modes, so motion of the middle race is directly transmitted to the outer spring plate (Fig.~\ref{fig:cross_section}(c)).

The \textit{inner spring plate} is permanently coupled to the output flange and rotates with the actuator output. It can also selectively engage with Inner Race--2 through a dog-tooth interface (Fig.~\ref{fig:cross_section}(c)).

A set of \textit{extension springs} connects the outer and inner spring plates. Four springs are arranged radially at equal angular intervals of $90^{\circ}$. The equivalent torsional stiffness is computed as follows.

Each spring has stiffness $k = 12.701$~N/mm and free length $l_0 = 12.8$~mm. The inner hooks are located at radius $r_1 = 25.32$~mm and the outer hooks at $r_2 = 37.87$~mm from the shaft center. During assembly, the springs are preloaded by approximately 1.75~mm, corresponding to a preload force of about 22~N per spring.

For a relative plate rotation $\beta$, the instantaneous spring length is
\begin{equation}
\label{eq:spring_length}
l(\beta) = \sqrt{r_1^2 + r_2^2 - 2\,r_1\,r_2\,\cos\beta}
\end{equation}

Resolving the spring force along the tangential direction, the total restoring torque from the four springs is
\begin{equation}
\label{eq:spring_torque}
\tau_\text{hub} = 4\,k\,r_1\,r_2\left(1 - \frac{l_0}{l(\beta)}\right)\sin\beta
\end{equation}

For small deflections, $\sin\beta \approx \beta$ and $l(\beta)$ is approximately constant at the preloaded length $l_p$. The hub stiffness becomes
\begin{equation}
K_s = \frac{d\tau_\text{hub}}{d\beta}\bigg|_{\beta \approx 0}
= 4\,k\,r_1\,r_2\left(1 - \frac{l_0}{l_p}\right)
\label{eq:hub_stiffness}
\end{equation}

where $l_p = l(0) + 1.75\,\text{mm}$ is the installed spring length. Substituting Table~\ref{tab:specs} yields $K_s \approx 5.8$\,Nm/rad, close to the measured $K_\text{SEA} = 5.57 \pm 0.02$\,Nm/rad (Section~\ref{sec:exp_stiffness}). The difference arises from structural compliance and friction not captured in the model. The moderate stiffness was chosen to make topology-dependent effects observable in this direct-drive prototype.

\subsection{Working of DTEA}
\label{sec:dtea_working}

\subsubsection{SEA Mode}
In the SEA operating mode, the solenoid is energized, pushing the middle race axially toward the actuator output. Consequently, the outer dog teeth of the middle race disengage from the outer race, while the inner dog teeth engage with Inner Race-1 attached to the rotating shaft.

Because the middle race is rigidly connected to the outer spring plate, this engagement causes the outer spring plate to rotate with the shaft. The inner spring plate remains disengaged from Inner Race-2 and is permanently coupled to the output flange.
The two plates are connected by springs, so torque must pass through the spring hub before reaching the output. The transmission path is
\begin{center}
Shaft $\rightarrow$ Inner Race--1 $\rightarrow$ Middle Race $\rightarrow$ Outer Spring Plate $\rightarrow$ Springs $\rightarrow$ Inner Spring Plate $\rightarrow$ Output.
\end{center}

This configuration places the elastic element between the motor and the output, forming a \textit{series elastic actuator} topology (Fig.~\ref{fig:cross_section}(a, top)).

\subsubsection{PEA Mode}

In the PEA operating mode, the solenoid is de-energized, pulling the middle race axially toward the motor. Consequently, the outer dog teeth of the middle race engage with the outer race, while the inner dog teeth disengage from Inner Race--1.

Because the outer spring plate is rigidly connected to the middle race, this axial motion shifts the outer spring plate toward the motor. The springs and inner spring plate translate axially as well, causing the inner spring plate to engage with Inner Race--2 through a dog-tooth interface. Inner Race--2 is rigidly attached to the shaft, creating a direct rigid connection between the shaft and the inner spring plate. This motion does not disengage the inner spring plate from the output flange.

Motor torque is therefore transmitted directly through the rigid path
\begin{center}
Shaft $\rightarrow$ Inner Race--2 $\rightarrow$ Inner Spring Plate $\rightarrow$ Output.
\end{center}

Simultaneously, the elastic element connects the output to the grounded housing through
\begin{center}
Outer Race (fixed) $\rightarrow$ Middle Race $\rightarrow$ Outer Spring Plate $\rightarrow$ Springs $\rightarrow$ Inner Spring Plate $\rightarrow$ Output.
\end{center}

Thus, the motor and output are rigidly coupled while the springs provide compliance between the output and the grounded housing, forming a \textit{parallel elastic actuator} (PEA) topology (Fig.~\ref{fig:cross_section}(a, bottom)).

\subsection{Switching Constraints}
\label{sec:switching_constraints}

Engagement succeeds at arbitrary relative angles due to the chamfered teeth. Disengagement requires sufficient unloading, as transmitted torque increases contact forces and friction at the interface. In the current prototype, the solenoid can overcome this friction for loads below approximately 1~Nm. Higher loads require briefly reducing transmitted torque. A higher-force voice-coil actuator is planned to extend switching under load~\cite{Grimmer2012}.

\begin{table}[h]
\centering
\caption{DTEA Prototype Specifications. Measured $K_\text{SEA}$ corresponds to the full deflection range. The linear-window value of 4.09~Nm/rad used in Section~\ref{sec:ol_stiffness} is reported in Section~\ref{sec:exp_stiffness}.}
\label{tab:specs}
\begin{tabular}{l l}
\hline
\textbf{Parameter} & \textbf{Value} \\
\hline
Motor & T-Motor U10 Plus KV100 \\
Torque constant $K_t$ & 0.083~Nm/A \\
Max.\ continuous current & 36~A (180~s) \\
Motor torque & 3.0~Nm (continuous) \\
Drive & Direct (no gearbox) \\
Motor Driver & ODrive Pro (FOC, 8~kHz loop) \\
\hline
Outer race ID / tooth count & 80~mm / 32 \\
Middle race OD, ID / teeth & 79~mm, 21~mm / $8 \times 2$ \\
Inner race OD / tooth count & 20~mm / 16 \\
Chamfer half-angle & $18^\circ$ (included $36^\circ$) \\
\hline
Spring rate (per spring) & 12.70~N/mm \\
Number of springs & 4 (at $90^\circ$ intervals) \\
Hook radii $r_1$, $r_2$ & 25.32~mm, 37.87~mm \\
Preload extension & $\approx$~1.75~mm \\
Analytical $K_s$ & 5.8~Nm/rad \\
Measured $K_\text{SEA}$ & 5.57 $\pm$ 0.02~Nm/rad \\
Measured $K_\text{PEA}$ & 8.54 $\pm$ 0.02~Nm/rad \\
\hline
Output encoder & AS5047P (14-bit absolute) \\
Solenoid force / stroke & 25~N / $\approx$~10~mm \\
Switching time & $<$~33.33~ms \\
Housing material & PLA (FDM) \\
\hline
\end{tabular}
\end{table}

\section{Experimental Validation}
\label{sec:experiments}

\begin{figure}[t]
    \centering
    \includegraphics[width=\columnwidth]{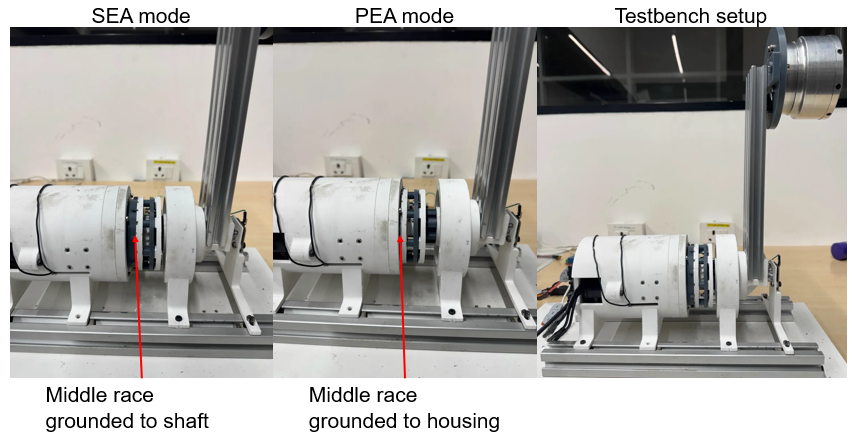}
    \caption{DTEA prototype and experimental testbench. \textit{(a)}~SEA mode, with the middle race engaged with Inner Race~1, connecting the spring to the shaft. \textit{(b)}~PEA mode, with the middle race engaged with the outer race, mechanically grounding the spring to the housing. The axial shift of the selector is visible between configurations. \textit{(c)}~Full testbench with 920\,g mass arm at 0.26\,m radius and AS5047P encoder at the output base.}
    \label{fig:testbench}
\end{figure}

Experiment~\ref{sec:exp_stiffness} uses a locked output fixture with the motor in torque control mode. Experiments~\ref{sec:exp_dynamic_switching}--\ref{sec:exp_disturbance} use a 920\,g mass arm at 0.26\,m radius (2.35\,Nm at horizontal). Motor-side position is measured by the ODrive's onboard encoder and output-side position by the AS5047P at 50\,Hz via Arduino serial link. Experiment~\ref{sec:exp_stiffness} uses torque control whereas experiments \ref{sec:exp_dynamic_switching}--\ref{sec:exp_disturbance} use position control on the ODrive. All data is logged at the controller's native rate (8\,kHz for motor-side, 50\,Hz for output-side). Fig.~\ref{fig:testbench} shows the experimental setup.

\subsection{Static-Stiffness Characterization}
\label{sec:exp_stiffness}

The effective mechanical stiffness of the actuator was experimentally evaluated in both Series Elastic Actuator (SEA) and Parallel Elastic Actuator (PEA) modes. The measured SEA stiffness is compared with the analytical stiffness derived in Section~\ref{sec:spring_hub}. In PEA mode, the output is directly coupled to the motor, which would ideally result in infinite stiffness. The finite stiffness observed experimentally indicates structural compliance within the transmission and couplings between the motor and output.

For the experiment, the actuator output shaft was rigidly fixed to ground using a mechanical fixture while the motor operated in torque control. The commanded torque $\tau_m$ followed a loading--unloading cycle: $0 \rightarrow +1 \rightarrow 0 \rightarrow -1 \rightarrow 0$\,Nm. Three cycles were recorded for each mode. Motor angle $\theta_m$ and torque $\tau_m$ were measured, and the effective stiffness was obtained as the slope of a linear least-squares fit of $\tau_m$ versus $\theta_m$.

\begin{figure}[t]
    \centering
    \includegraphics[width=\columnwidth]{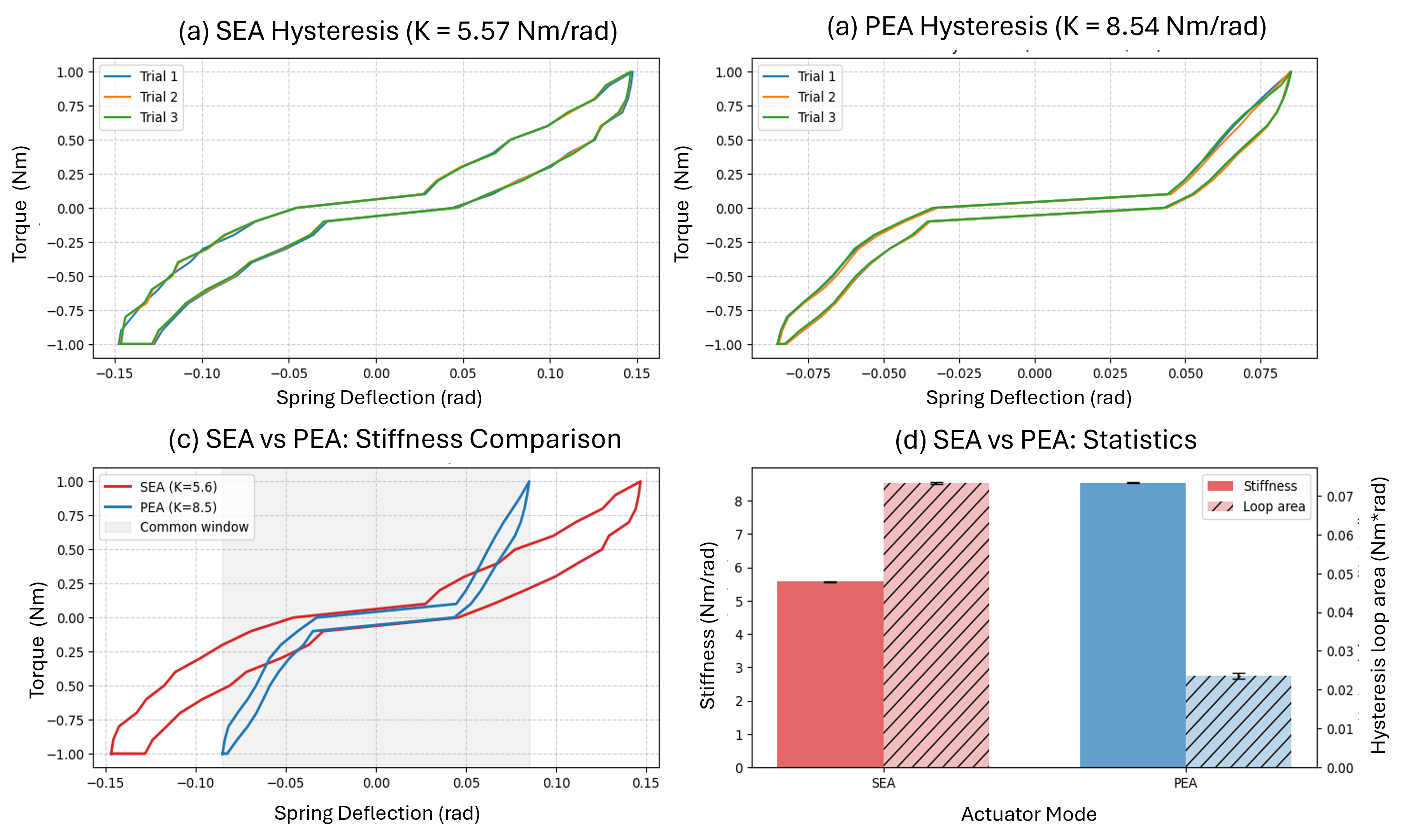}
    \caption{Stiffness characterization. \textit{(a)}~SEA hysteresis loops (3 trials). \textit{(b)}~PEA hysteresis loops (3 trials). \textit{(c)}~Overlay with common deflection window shaded. \textit{(d)}~Stiffness and hysteresis loop area comparison.}
    \label{fig:stiffness}
\end{figure}

The resulting torque--deflection curves are shown in Fig.~\ref{fig:stiffness}(a)--(b). Both modes exhibit approximately linear behavior within the common deflection window shown shaded in Fig.~\ref{fig:stiffness}(c), used for the $2.08\times$ ratio in Section~\ref{sec:ol_stiffness}. The full-range value $K_{\text{SEA}} = 5.57$\,Nm/rad applies at larger deflections such as those in Section~\ref{sec:exp_dynamic_switching}. The estimated stiffness values are $K_{\text{SEA}} = 5.57 \pm 0.02$\,Nm/rad and $K_{\text{PEA}} = 8.54 \pm 0.02$\,Nm/rad (mean $\pm$ s.d., $n=3$), while the analytical prediction is $K_{\text{SEA}} = 5.8$\,Nm/rad. The slightly lower experimental SEA stiffness is attributed to compliance in the 3D-printed structural components not captured by the analytical model.

The PEA mode exhibits higher stiffness than SEA, consistent with the theoretical relation $K_{\text{PEA}}/K_{\text{SEA}} = 1 + (K_{struct}/K_s)$ (Eq.~\ref{eq:ol_ratio}). Hysteresis, quantified by the loading--unloading loop area, is reduced by $67.7\%$ in PEA mode compared to SEA ($0.024$ vs.\ $0.073$\,Nm$\cdot$rad), as summarized in Fig.~\ref{fig:stiffness}(d).

\begin{figure}[t]
    \centering
    \includegraphics[width=0.85\columnwidth]{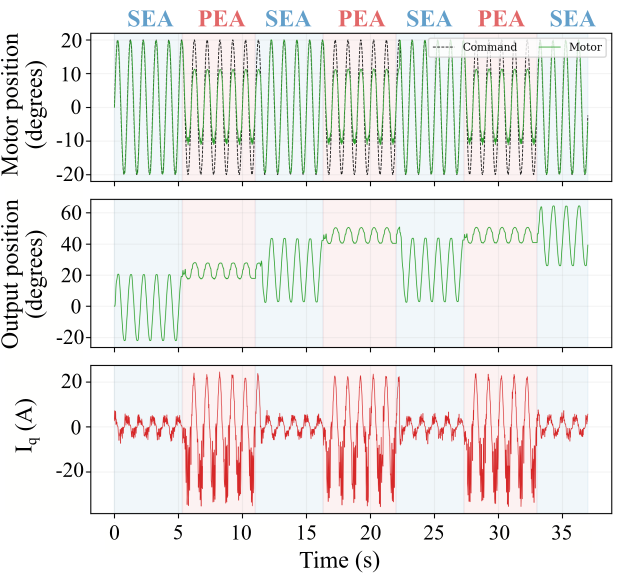}
    \caption{Real-time topology switching during 1\,Hz sinusoidal tracking. Blue/red shading indicates SEA/PEA mode. \textit{Top.} Motor position vs.\ command. \textit{Middle.} Output position. \textit{Bottom.} Motor current $I_q$.}
    \label{fig:dynamic_switching}
\end{figure}

\subsection{Dynamic Switching under cyclic loading}
\label{sec:exp_dynamic_switching}

The objective of this experiment is to demonstrate dynamic switching of the DTEA between SEA and PEA operating modes. A sinusoidal motor target position $\theta_m^{tar}$ with frequency $1\,\text{Hz}$ and amplitude $\pm 20^\circ$ is commanded. The motor is controlled using a proportional position controller
\[
\tau_m = K_p (\theta_m^{tar} - \theta_m)
\]
where $\theta_m$ is the measured motor position and $K_p = 40$. The actuator switches between SEA and PEA modes every 5~s while operating. An encoder mounted at the output measures the output position $\theta_o$, and the motor current $I_q$ is also recorded.

Fig.~\ref{fig:dynamic_switching} shows the results. The results illustrate distinct actuator behavior in the two modes despite using the same controller gain. In SEA mode the motor closely tracks the commanded motion, whereas larger tracking errors occur in PEA mode due to the restoring effect of the parallel spring. The output motion reflects this behavior: in SEA mode the output follows the commanded oscillation, while in PEA mode the oscillation amplitude is reduced because the spring resists changes in equilibrium position.

The equilibrium position of the PEA oscillation shifts after each transition since switching occurs at arbitrary output positions. Maintaining a constant equilibrium would require switching at the oscillation center, which was not enforced as the objective was to verify reliable switching.

The motor current plot shows higher peak currents in PEA mode because the motor must work against the parallel spring to take the output away from the equilibrium position. Overall, the results demonstrate successful dynamic switching of the DTEA between SEA and PEA topologies, with each configuration exhibiting its expected dynamic behavior.

\subsection{Switching Time Calculation}
\label{sec:exp_switching}

\begin{figure}[t]
    \centering
    \includegraphics[width=0.9\columnwidth]{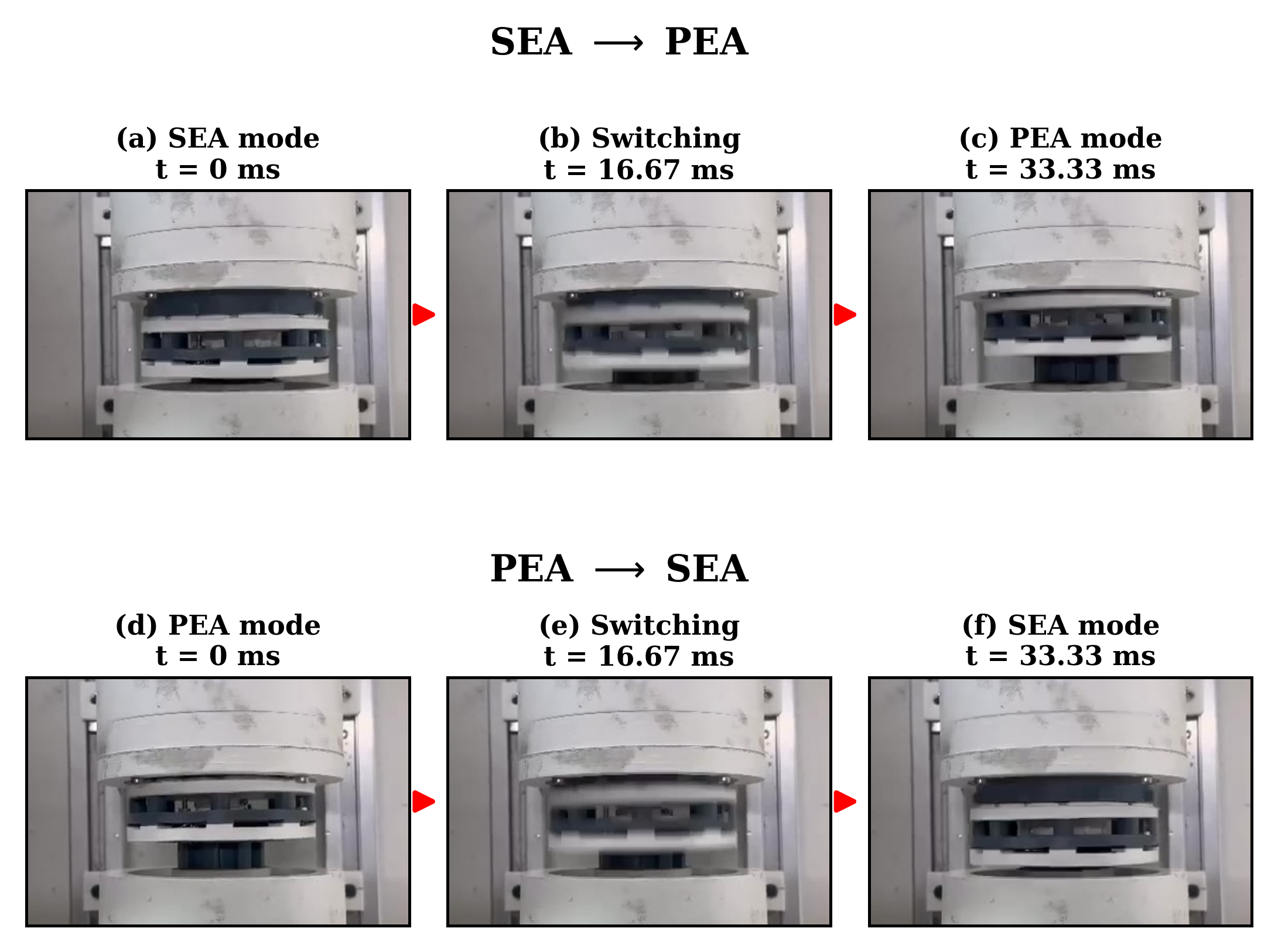}
    \caption{Switching time characterization via high-speed video at 60\,fps (16.67\,ms per frame). \textit{Top row:} SEA$\to$PEA transition. \textit{Bottom row:} PEA$\to$SEA transition. In each row, the first frame shows the middle race at rest in the initial topology, the second frame captures the race mid-travel, and the third frame shows the race at rest in the final topology, bounding the switching time to $<$\,33.33\,ms.}
    \label{fig:switching_time}
\end{figure}

This experiment quantifies the switching time of the topology transition. The mechanism was recorded using an iPhone camera at 60\,fps (16.67\,ms per frame) and frames were extracted at the native capture rate without interpolation. Fig.~\ref{fig:switching_time} shows three consecutive frames for each transition direction. The first frame shows the middle race at rest in the original topology. The second frame, captured 16.67\,ms later, shows the race in transit between engagement positions. The third frame, at 33.33\,ms, shows the race at rest in the new topology. This sequence was consistent across all observed transitions in both directions, bounding the switching time to $<$\,33.33\,ms. This measurement was performed under static no-load conditions. In dynamic operation, the transient visible in Fig.~\ref{fig:dynamic_switching} arises from tooth engagement at an arbitrary relative angle, producing a brief compliance mismatch before the new topology is established.

\subsection{Disturbance Rejection in position control mode}
\label{sec:exp_disturbance}

The objective of this experiment is to evaluate the actuator response to unexpected external disturbances in both SEA and PEA operating modes. The motor was commanded to hold a fixed horizontal position using position control with gain $K_p = 30$. External disturbances were introduced by manually striking the mass arm with a rubber mallet to generate impulsive impacts. The mallet force was not instrumented and the applied energy was not controlled to a fixed value across trials. Six impacts were recorded in SEA mode and five in PEA mode. Peak deflection was the maximum angular deviation after impact. Settling time was measured until the position remained within $\pm 0.5^\circ$ of the command, five times the encoder noise floor ($\pm0.1^\circ$).

\begin{figure}[t]
    \centering
    \includegraphics[width=0.95\columnwidth]{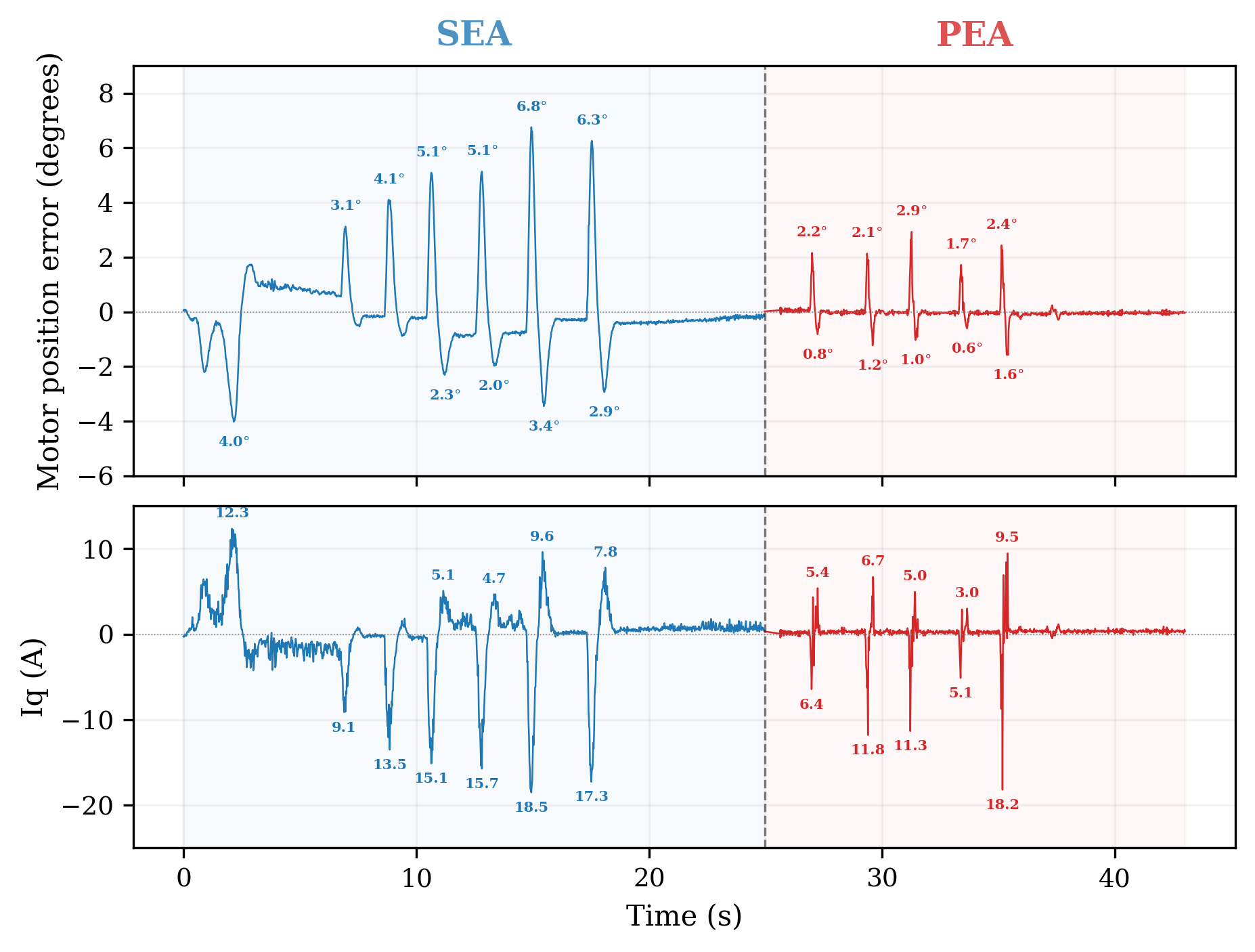}
    \caption{Disturbance rejection under position control 
    ($K_p = 30$). \textit{Top.} Motor position error for 6 SEA 
    (blue) and 5 PEA (red) mallet impacts with per-trial peak 
    deflections annotated. \textit{Bottom.} Corresponding motor 
    current $I_q$. SEA settling is dominated by oscillation between the motor and load through the spring.}
    \label{fig:disturbance}
\end{figure}

Fig.~\ref{fig:disturbance} shows the impulse responses for both modes. The initial negative deflection in the SEA trace corresponds to the arm settling at the commanded position before disturbance and is excluded from analysis. Averaged across trials, the mean peak deflection in SEA mode is $2.26\times$ larger than in PEA mode ($5.2^\circ$ vs.\ $2.3^\circ$), while the mean settling time is $3.45\times$ longer ($1380$\,ms vs.\ $400$\,ms).

Under position control the motor behaves as a virtual spring--damper system. In SEA mode the physical spring is connected in series with this virtual stiffness, whereas in PEA mode it is connected in parallel. The series configuration reduces effective stiffness, while the parallel configuration increases it. Consequently, disturbances produce larger and longer oscillations in SEA mode than in PEA mode, consistent with the expected dynamics of SEA and PEA architectures.

\section{Discussion}
\label{sec:discussion}

The experimental results confirm that real-time topology 
switching produces measurably different static and 
dynamic behavior on the same hardware under identical 
controller gains. The observed differences in stiffness, 
disturbance response, and motor current arise from the 
elastic grounding path rather than from controller retuning 
or stiffness adjustment.

The disturbance rejection results in Section IV-D and Fig. 8 match the structural difference described in Section~\ref{sec:dynamics}. In SEA mode, the motor and load 
are coupled through the spring, forming a two-mass system 
as described in~\cite{Pratt1995}. This allows relative motion 
between motor and load and produces oscillation following 
disturbance. A motor-side position controller cannot fully 
suppress this internal mode~\cite{Calanca2016}. In PEA mode, 
the motor and load are rigidly coupled and the spring is 
grounded to the housing, eliminating this internal resonance. 
The larger peak deflection and longer settling time observed 
in SEA, and the reduced oscillation in PEA, follow directly 
from this structural difference. All disturbance trials were 
performed with identical position gains.

The open-loop stiffness measurements in 
Section~\ref{sec:exp_stiffness}, Fig.~\ref{fig:stiffness}, 
and Table~\ref{tab:specs} show a $2.08\times$ 
stiffness ratio between modes. In SEA mode, apparent 
stiffness is bounded by $K_s$. In PEA mode, apparent 
stiffness includes both $K_s$ and the structural load path. 
The finite value of $K_\text{struct}$ is due to compliance 
in the 3D-printed components. A machined metal implementation 
would increase structural stiffness and increase the 
separation between SEA and PEA behavior.

Several limitations of the current prototype should be 
noted. First, disengagement requires transmitted torque 
below approximately $1$~Nm due to friction at the dog-tooth 
interface (Section~\ref{sec:switching_constraints}). Engagement is not 
torque-dependent, but switching under load is currently 
limited. The switching time bound applies under static no-load conditions. A higher-force voice-coil actuator is under development to extend this operating range. Second, the 
3D-printed construction limits torque capacity and increases 
structural compliance (Table~\ref{tab:specs}). 
Third, the disturbance dataset in 
Section~\ref{sec:exp_disturbance} contains a limited number 
of impacts, although the mode-dependent trends are consistent 
across trials. Finally, energy reduction, which motivates 
PEA adoption in prior work~\cite{Verstraten2016,Beckerle2017}, 
was not evaluated at the task level. The reported 
$4.93\times$ current ratio in 
Section~\ref{sec:exp_dynamic_switching} reflects topology-dependent load sharing during sinusoidal tracking rather 
than application-specific motion. Task-level energy 
evaluation remains future work.

\section{Conclusion}
\label{sec:conclusion}

This paper presented the Dual-Topology Elastic Actuator, 
which to our knowledge is the first actuator capable of 
real-time switching between series and parallel 
elastic configurations. A three-race dog selector mechanism 
redirects the connection point of a single radial spring hub 
between the motor shaft and the actuator housing via 
solenoid-driven axial translation, achieving topology 
transitions in ${<}\,33.33$\,ms.
Experiments on the same hardware confirmed distinct stiffness, disturbance, and current characteristics between topologies. Open-loop stiffness differs by $2.08\times$ 
between modes with 67.7\% lower hysteresis in PEA. 
Disturbance rejection tests showed $2.26\times$ lower peak 
deflection and $3.45\times$ faster settling in PEA, with 
SEA settling dominated by oscillation between the motor 
and load through the spring. Real-time switching during 
1\,Hz tracking demonstrated a $4.93\times$ reduction in 
motor current upon transitioning from SEA to PEA, and all 
324 switching events completed within ${<}\,33.33$\,ms.
Future work will focus on a machined metal prototype with higher torque capacity, switching under load using voice-coil actuation, closed-loop frequency characterization, and task-level energy validation.

\bibliographystyle{IEEEtran}
\bibliography{references}

\end{document}